\documentclass[letterpaper, 10 pt, journal, twoside]{IEEEtran}

\usepackage{amsmath,amsfonts,amssymb}
\usepackage{array}
\usepackage[caption=false,font=normalsize,labelfont=sf,textfont=sf]{subfig}
\usepackage{textcomp}
\usepackage{stfloats}
\usepackage{url}
\usepackage{verbatim}
\usepackage{graphicx}
\usepackage{cite}
\usepackage{comment}
\usepackage[hidelinks]{hyperref}
\hypersetup{
   linkcolor=blue,
   breaklinks=true,   % splits links across lines
   colorlinks=true,   % displays links as colored text instead of blocks
   citecolor=blue,
   urlcolor=blue
}
\usepackage{siunitx,booktabs}
\usepackage{multirow}
\usepackage[table,xcdraw]{xcolor}
\usepackage{colortbl}
\usepackage{hhline}
\usepackage[font=footnotesize,labelfont=bf,tableposition=top]{caption}
\usepackage{soul}
\usepackage{graphicx}

\newcommand{\norm}[1]{\left\lVert#1\right\rVert}
\definecolor{hidden-draw}{RGB}{205, 44, 36}
\definecolor{lightgray}{gray}{0.9}
\definecolor{lightblue}{RGB}{173, 216, 230}

\usepackage[ruled,vlined,linesnumbered]{algorithm2e}

\SetCommentSty{mycommfont}
\SetKwComment{Comment}{$\triangleright$\ }{}
\let\oldnl\nl%
\newcommand{\nonl}{\renewcommand{\nl}{\let\nl\oldnl}}%

\usepackage{titlesec}
\titlespacing*{\subsection}{0pt}{0.1cm}{0.1cm}

\usepackage{tcolorbox}
\usepackage[T1]{fontenc}

\begin{document}

% Paper sections
\title{\textbf{SIMPNet}: \underline{S}patial-\underline{I}nformed \underline{M}otion \\ \underline{P}lanning \underline{Net}work}

\author{Davood Soleymanzadeh$^{1}$, Xiao Liang$^{2}$, and Minghui Zheng$^{1}$
\thanks{Manuscript received: August, 22, 2024; Revised October, 19, 2024; Accepted January, 22, 2025.}
        % <-this % stops a space
\thanks{This paper was recommended for publication by Editor Aniket Bera upon evaluation of the Associate Editor and Reviewers' comments. This work was supported by
USA National Science Foundation under Grant 2026533/2422826. (\textit{Corresponding authors: Minghui Zheng; Xiao Liang.})}% <-this % stops a space
\thanks{$^{1}$Davood Soleymanzadeh and Minghui Zheng are with the J. Mike Walker '66 Department of Mechanical Engineering, Texas A\&M University, College Station, TX 77843, USA (\tt\footnotesize e-mail: davoodso@tamu.edu; mhzheng@tamu.edu).}% <-this % stops a space
\thanks{$^{2}$Xiao Liang is with the Zachry Department of Civil and Environmental
Engineering, Texas A\&M University, College Station, TX 77843 USA (\tt\footnotesize e-mail:
xliang@tamu.edu).}
\thanks{Supplementary materials are available via this \href{https://zh.engr.tamu.edu/wp-content/uploads/sites/310/2024/12/Revised-SIMPNet-II-Demo.mp4}{link}.}
\thanks{Digital Object Identifier (DOI): see top of this page.}}
% Paper headers
\markboth{IEEE Robotics and Automation Letters. Preprint Version. January, 2025}
{Soleymanzadeh \MakeLowercase{\textit{et al.}}: SIMPNet}
% Use only for final RAL version}
\maketitle
\begin{abstract}
Current robotic manipulators require fast and efficient motion-planning algorithms to operate in cluttered environments. State-of-the-art sampling-based motion planners struggle to scale to high-dimensional configuration spaces and are inefficient in complex environments. This inefficiency arises because these planners utilize either uniform or hand-crafted sampling heuristics within the configuration space. To address these challenges, we present the Spatial-informed Motion Planning Network (SIMPNet). SIMPNet consists of a stochastic graph neural network (GNN)-based sampling heuristic for informed sampling within the configuration space. The sampling heuristic of SIMPNet encodes the workspace embedding into the configuration space through a cross-attention mechanism. It encodes the manipulator's kinematic structure into a graph, which is used to generate informed samples within the framework of sampling-based motion planning algorithms. We have evaluated the performance of SIMPNet using a UR5e robotic manipulator operating within simple and complex workspaces, comparing it against baseline state-of-the-art motion planners. The evaluation results show the effectiveness and advantages of the proposed planner compared to the baseline planners. Project website: \href{https://davoodsz.github.io/simpnet/}{https://davoodsz.github.io/simpnet/}
\end{abstract}

\begin{IEEEkeywords}
Motion and Path Planning, Imitation Learning, Graph Neural Networks (GNNs), Attention Mechanism.
\end{IEEEkeywords}
\section{Introduction} \label{sec: intro}
\IEEEPARstart{M}{otion} planning is a critical component within the robotic autonomy stack, enabling robotic manipulators to achieve task-specific goals \cite{orthey2023sampling}. For a robotic manipulator, this process involves finding a feasible collision-free path between a pre-defined start and goal within its configuration space. Over the past few years, a diverse collection of motion planning algorithms has been developed to address the motion planning problem \cite{orthey2023sampling}. Among motion planning paradigms, probabilistic complete sampling-based motion planners are effective for robotic manipulators operating in complex environments, allowing them to navigate safely and efficiently \cite{lavalle2006planning}. 

Sampling-based motion planners rely on three algorithmic primitives: sampling within high-dimensional configuration spaces, steering towards these sampled configurations, and checking collisions for such connections \cite{lavalle2006planning}.  These planners iteratively build a tree by connecting collision-free samples drawn from their underlying sampling distribution. For example, Rapidly-Exploring Random Trees (RRTs) \cite{lavalle2006planning} uniformly sample the configuration space, a method that becomes computationally intensive within high-dimensional configuration spaces. To address this, more recent advanced sampling-based motion planners utilize hand-crafted heuristics to modify the sampling distribution, biasing it towards areas that likely reduce planning cost and increase success likelihood \cite{gammell2014informed}. However, challenges like developing an initial path to guide these heuristics and the inherent complexity of high-dimensional configuration spaces can hinder the effectiveness of these planners \cite{qureshi2018deeply}. 

\begin{figure}[t] 
\centering
\includegraphics[width=3 in]{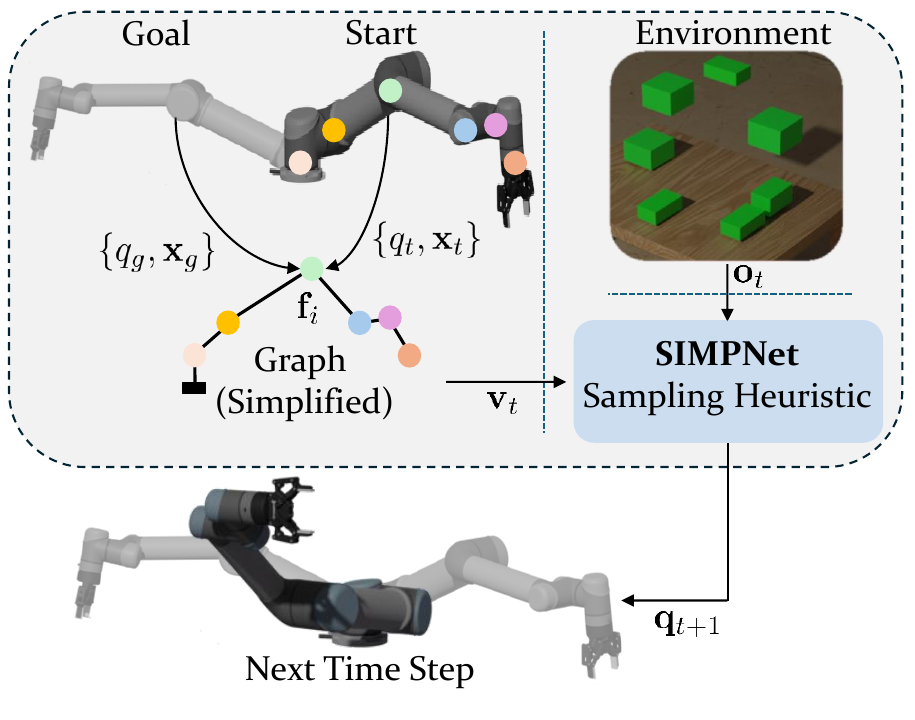}
\caption{An example of SIMPNet planning in a complex environment. The proposed sampling heuristic (Figure \ref{fig_2- simpnet-framework}) within the SIMPNet framework leverages graph and graph neural network frameworks to retain the spatial relationships within the configuration space and the kinematic chain of the robotic manipulator. This approach allows for informed sampling within the framework of sampling-based motion planning algorithms. $q_t$, and $\mathbf{x}_t$ are current joint angle and Cartesian 3D workspace position respectively. $q_g$, and $\mathbf{x}_g$ are the goal joint angle and Cartesian 3D workspace position respectively. $\mathbf{f}_i$ denotes node features, $\mathbf{v}_t$ denotes the concatenation of all nodes' features, and $\mathbf{o}_t$ denotes environment embedding. $\mathbf{q}_{t+1}$ is the next time step sampled manipulator's configuration.}
\label{fig_1- simpnet-example}
\end{figure}

Recently, learning-based motion planning methods have been developed to improve the sampling heuristic of traditional sampling-based planners. These methods use networks trained on successful paths to capture the similarity between planning problems and encode the underlying structure of the configuration space. As a result, they generate an informed sample based on the learned sampling heuristic \cite{johnson2023learning}. However, many of these learning-based planners do not fully account for the spatial relations within the configuration space and workspaces, or the sequential structure inherent within the motion planning problem \cite{zang2022robot}.

We introduce SIMPNet, a novel motion planner that features a stochastic neural sampling heuristic based on a message-passing neural network architecture \cite{battaglia2018relational}. This approach enables SIMPNet to learn motion policies and generate samples that are informed by the spatial relationships within the configuration space. By integrating relational information from the robotic manipulator's kinematic structure into a graph and applying a cross-attention mechanism \cite{vaswani2017attention} to merge workspace embeddings with configuration space, SIMPNet facilitates spatially-aware sampling. Combined with a bi-directional planning algorithm \cite{qureshi2019motion}, this method reduces planning time and enhances the success rate, thus outperforming existing motion planning algorithms like MPNet \cite{qureshi2019motion}.

The contributions of this work include:
\begin{itemize}
    \item We construct a graph that implicitly represents the kinematic chain of the robotic manipulator, providing a spatial representation of its movements within the configuration space. We use a message-passing neural network structure \cite{battaglia2018relational}, to perform message-passing on the constructed graph. This process ensures the sampling heuristic within the SIMPNet framework is spatially aware of the configuration space, thereby facilitating informed sample generation.
    
    \item We integrate a cross-attention mechanism into our sampling heuristic, which effectively incorporates workspace embeddings into the configuration space \cite{zhang2022learning}. The cross-attention module addresses challenges posed by the differing dimensionalities of the configuration space and workspace, allowing for efficient integration of information between these two environments.
    
    \item We evaluate the performance of SIMPNet across various workspaces, empirically demonstrating that it outperforms other benchmark motion planning algorithms in terms of planning time and success rate.
\end{itemize}

The paper is structured as follows. Section \ref{sec: related work} reviews related work in sampling-based motion planning. Section \ref{sec: GNN-based planning} introduces the SIMPNet structure. Section \ref{sec: results and discussion}  presents a performance evaluation of our proposed planner, and compares it against benchmark motion planners. Finally, section \ref{sec: conclusion} concludes the paper.

\section{Related Work} \label{sec: related work}
This section briefly reviews recent literature on sampling-based motion planning for robotic manipulators.

\vspace{0.1cm}
\noindent
\textbf{Sampling-based motion planning (SBMP)}: Sampling-based motion planners, such as RRTs \cite{lavalle2006planning} have emerged as a practical solution for motion planning of robotic manipulators. SBMP algorithms are probabilistic complete, meaning the probability of finding a feasible path, if one exists, approaches one as the number of samples increases to infinity \cite{lavalle2006planning}. SMBP algorithms use sampling techniques to generate rapidly-exploring trees (single-query rapidly-exploring trees) \cite{lavalle2006planning} within the manipulator's configuration space. These planners operate directly within the continuous configuration space without requiring precise models of collision and collision-free spaces \cite{lavalle2006planning}.

RRTs \cite{lavalle2006planning} sample the configuration space uniformly to construct a tree between the start and goal configuration for path planning. However, RRTs struggle to find the optimal path. RRT* (i.e., optimal RRT), an extension of RRTs, aims to achieve asymptotic optimality \cite{karaman2011sampling}. RRT* employs rewiring steps within the constructed tree to find a sub-optimal path, necessitating more samples. As the manipulator's degrees of freedom (DOF) increase, so does the computational complexity. RRTs and RRT*  utilize a uniform sampling heuristic to construct a tree implicitly representing the configuration space. However, this uniform sampling heuristic struggles to scale to high-dimensional configuration spaces, leading to higher computational complexity and longer planning time. More advanced SBMPs such as Informed-RRT* \cite{gammell2014informed} and Batch-Informed Trees (BIT*) \cite{gammell2015batch} employ informed sampling heuristics that focus on areas likely to contain optimal paths, thereby reducing computational complexity and planning time. These hand-crafted sampling heuristics decrease the planner's computation complexity and planning time by directing sampling towards regions with potential optimal paths. Nevertheless, crafting these informed sampling heuristics is not trivial, particularly for high-dimensional configuration spaces \cite{qureshi2020motion}. The major drawbacks of SBMPs are sample inefficiency and expensive collision checking \cite{yu2024efficient}.

\vspace{0.1cm}
\noindent
\textbf{Neural-based SBMPs}: Neural-based SBMPs utilize deep learning frameworks - known for fast inference, inductive bias, and the capability to encode the multi-modal structure within datasets - to replace or enhance algorithmic primitives of SBMPs. For the sampling heuristic, deep learning modules have been utilized to either implicitly learn the heuristic for generating informed samples, or explicitly generate samples. Deep generative models \cite{sohn2015learning} are particularly popular for learning the underlying sampling heuristic for a specific motion planning problem \cite{johnson2023learning}. Additionally, the rapid inference capabilities of multi-layer perceptrons (MLPs) are employed as explicit informed sampling heuristic in motion planning \cite{qureshi2019motion}. However, many of these planners struggle to adequately account for the spatial and temporal dependencies inherent in motion planning problems.

\vspace{0.1cm}
\noindent
\textbf{Graph neural networks (GNNs)-based SBMPs}: GNNs-based SBMPs leverage graphs, which are powerful for representing both structured and unstructured data, and for encoding temporal and spatial relationships within them \cite{battaglia2018relational}. Graph neural networks (GNNs), which operate on graphs, offer a structured representation that effectively encodes spatial correlations within a dataset \cite{battaglia2018relational}. This capability is useful for encoding spatial dependencies within a robotic manipulator's configuration space. Researchers \cite{yu2021reducing} have utilized GNNs to enhance edge evaluation in planning trees, thereby reducing the need for expensive edge evaluation and collision checks. However, a major drawback of these approaches is that the initial geometric graphs are often generated randomly, without consideration for the underlying structure of the configuration space.

In contrast to these works, our framework, SIMPNet, uses a stochastic neural network built on a message-passing neural network framework, serving as a spatially informed sample generator. The aim of SIMPNet's sampling primitive is to learn an informed sampling heuristic from a set of observed sub-optimal paths. This innovative learning approach is designed to improve the success rate and reduce the planning time required by the motion planning algorithm.

\section{Spatial-Informed Motion Planning Network (SIMPNet)} \label{sec: GNN-based planning}
\begin{figure*}[htbp] 
\centering
\includegraphics[width=\textwidth]{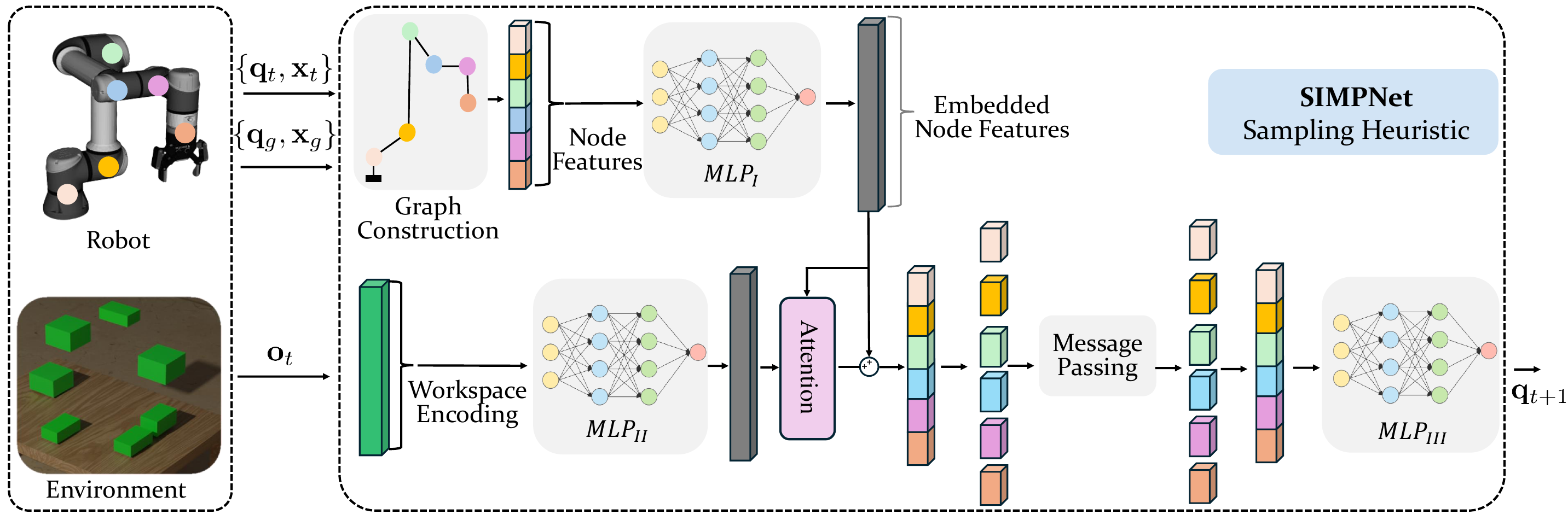}
\caption{Schematic of spatial-informed sampling heuristic within SIMPNet. Utilizing the current time step planning information from the workspace and configuration space, this sampling heuristic generates the next time step configuration, moving towards the goal of the motion planning problem. $q_t$, and $\mathbf{x}_t$ are the current joint angle and Cartesian 3D workspace position respectively. $q_g$, and $\mathbf{x}_g$ are the goal joint angle and Cartesian 3D workspace position respectively. $\mathbf{o}_t$, and $\mathbf{q}_{t+1}$ are the environment embedding, and the next time step sampled manipulator's configuration respectively.}
\label{fig_2- simpnet-framework}
\end{figure*}

We introduce an informed sampling heuristic within the SIMPNet framework that uses current timestep planning information from configuration space and workspace to generate the next timestep configuration in a motion planning problem. Figure \ref{fig_2- simpnet-framework} illustrates the schematic of the proposed sampling heuristic within the SIMPNet structure.

\subsection{Motion Planning Definition}
Let $\mathcal{C} \in \mathbb{R}^n$ represent the configuration space of a robotic manipulator where $n$ denotes the robot's degree of freedom (DOF). The configuration space comprises two spaces: the obstacle space ($\mathcal{C}_{obs} \subset \mathcal{C}$), and the free space ($\mathcal{C}_{free}=\mathcal{C}\backslash \mathcal{C}_{obs}$). Given a start configuration ($\mathbf{q}_{start} \in \mathcal{C}_{free}$), and a goal configuration ($\mathbf{q}_{goal} \in \mathcal{C}_{free}$), the motion planning problem involves finding a feasible path ($\mathbf{q}=\{\mathbf{q}_{start}, \cdots, \mathbf{q}_{goal}\}$) between the start and goal configuration such that the entire path lies within the free configuration space. 

\subsection{Robotic Manipulator's Graph Representation}
We model the kinematic structure of the robotic manipulator using an undirected graph. In the graph, $G = (V, E, F)$, $V$ represents the set of nodes, $E$ the set of edges, and $F$ the node features. An edge $e_{ij} \in E$ connects node $v_i \in V$ and $v_j \in V$, if they are connected. Each node feature vector $\mathbf{f}_i \in F$ is a $d$-dimensional vector ($\in \mathbb{R}^d$) associated with the corresponding node $v_i$. The neighborhood of a node $v_i$, denoted as $\mathcal{N}(i) = \{j|e_{ij} \in E\}$, represents the nodes directly connected to the given node $v_i$.

\begin{figure}[htbp] 
\centering
\includegraphics[width=3in]{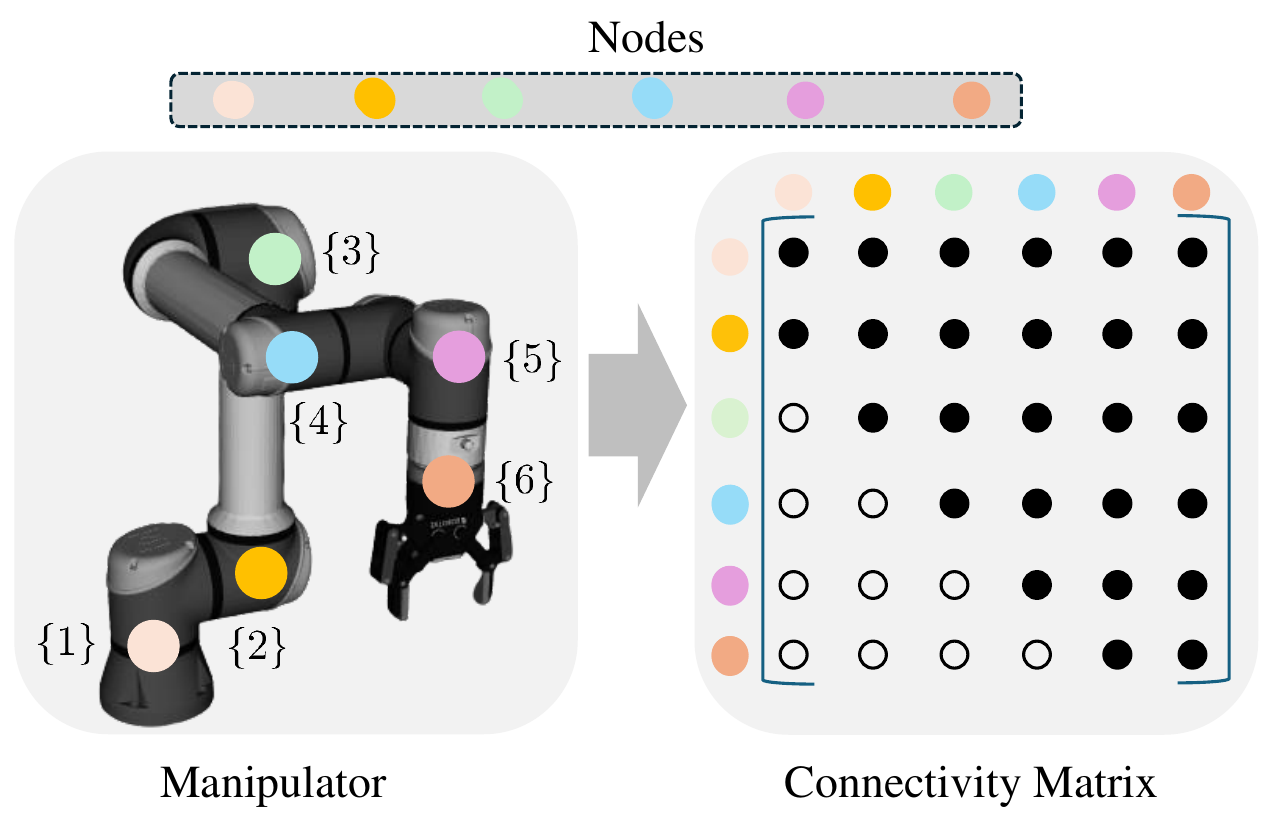}
\caption{Illustration of the constructed graph for the 6-DoF UR5e robotic manipulator. This graph represents the relational geometry and kinematics chain of the robotic manipulator. Each joint is considered a node, and edges are defined to reflect the kinematic connections between joints, mapping the robotic manipulator's structure into the graph.}
\label{fig_3- cobot-graph}
\vspace{-0.1cm}
\end{figure}

In our model, the joints of the robotic manipulator are represented as nodes in a graph, where the edges represent the topological and kinematic relationships between these nodes. The edges are explicitly defined based on the kinematic structure of the manipulator, with each node connected to the subsequent nodes following the kinematic chain of the robotic manipulator from the base joint toward the end-effector. The node features $(\mathbf{f}_i \in \mathbb{R}^{11})$ are selected to encapsulate relevant information, including the spatial information of both workspace and configuration space, kinematic information of the manipulator, and essential elements of motion planning as follows: 
\begin{equation} \label{eq - node_features}
\begin{split}
    \mathbf{f}_i = [&\mathbf{x}_t, \mathbf{x}_{goal}, |\mathbf{x}_{goal} - \mathbf{x}_t|, \norm{\mathbf{x}_{goal} - \mathbf{x}_t}_2, q_t, q_{goal}, \\
    & |q_t - q_{goal}|],
\end{split}
\end{equation}
where $\mathbf{x}_t~\in \mathbb{R}^3$, $\mathbf{x}_{goal}~\in \mathbb{R}^3$ are the current and goal positions in Cartesian 3D coordinates for each node, derived using forward kinematics (FK). Similarly, $q_t~\in \mathbb{R}$, and $q_{goal}~\in \mathbb{R}$ denote the current and target joint angles respectively. Figure \ref{fig_3- cobot-graph} illustrates the graph representation of our robotic manipulator.

\subsection{SIMPNet Sampling Heuristic Components}
Building on the graph constructed in the previous section, we now detail each component of the proposed sampling heuristic.

\noindent
\textbf{Embedding Workspace and Node Features}: Node features, as defined in eq. \ref{eq - node_features}, are concatenated to form a single embedding vector that implicitly represents the configuration space at time step $t$. Dimensions (length, height, width) and 3D coordinates of each obstacle's center are similarly concatenated to create the environment embedding at the same time step. Two distinct deep multi-layer perceptrons (MLPs) then transform the configuration space and the workspace embeddings into same-size embedding vectors as follows:
\begin{equation} \label{eq - taskspace}
\begin{aligned}
    \mathbf{v} = MLP_{I}\big(\parallel_{i=1}^{n_v}\mathbf{f}_i \parallel\big),
\end{aligned}
\end{equation}
\begin{equation} \label{eq -workspace}
\begin{aligned}
    \mathbf{o} = MLP_{II}\big(\parallel_{j=1}^{n_o}[x_j, y_j, z_j, L_j, W_j, H_j] \parallel \big),
\end{aligned}
\end{equation}
where $\mathbf{f}_i~\in \mathbb{R}^{11}$ represents the feature of $i$-th node as defined in eq. \ref{eq - node_features}, $n_v$ denotes the total number of nodes in the graph, and $\mathbf{v} \in \mathbb{R}^{d_z}$ denotes the configuration space embedding. $(x_j, y_j, z_j)~\in \mathbb{R}^3$ and $(L_j, W_j, H_j)~\in \mathbb{R}^3$  denote the Cartesian coordinate, and the dimensions (length, width, height) of the $j$-th obstacle within the workspace, respectively, $n_o$ denotes the number of obstacles, and $\mathbf{o}\in \mathbb{R}^{d_z}$ denotes the workspace embedding. The $\parallel$ operator represents the concatenation operator.

\noindent
\textbf{Integrating Workspace embedding into Configuration Space}: We utilize a cross-attention mechanism to integrate workspace embeddings, i.e., obstacles, into the configuration space. The attention mechanism, a key component of the Transformer model, helps encode inter-dependencies among elements \cite{vaswani2017attention}. The formulation of the attention mechanism is as follows:
\begin{equation} \label{eq - attention-mechanism}
\begin{aligned}
    \text{Attention}(\mathbf{q}, \mathbf{k}, \mathbf{v}) = \text{softmax}(\frac{\mathbf{q}\mathbf{k}^T}{\sqrt{d_k}})\mathbf{v},
\end{aligned}
\end{equation}
where $\mathbf{q}~\in \mathbb{R}^{d_k}$, $\mathbf{k}~\in \mathbb{R}^{d_k}$, and $\mathbf{v}~\in \mathbb{R}^{d_k}$ are query, key, and value vectors respectively, with $d_k$ indicating the dimensionality of the key vector. This attention mechanism integrates workspace embeddings into the configuration space in our research.
\begin{equation} \label{eq -cross-attention}
\begin{aligned}
    \mathbf{v} = \text{Cross-attention}(\mathbf{q}_v, \mathbf{k}_o, \mathbf{v}_o) + \mathbf{v},
\end{aligned}
\end{equation}
where $\mathbf{q}_v~\in \mathbb{R}^{d_z}$, $\mathbf{k}_o~\in \mathbb{R}^{d_z}$, and $\mathbf{v}_o~\in \mathbb{R}^{d_z}$ are query, key and value vectors respectively. These vectors are derived using one-layer MLPs.

\noindent
\textbf{Kinematics Aware Message Passing}: Following the integration of workspace information into the configuration space, and the construction of the graph, we aggregate node features to predict node-level (joint-level) properties \cite{battaglia2018relational}. To this end, we utilize a message-passing neural network (MPNN) \cite{battaglia2018relational} which consists of three steps for performing message passing to produce a new graph with the same structure and connectivity but updated node features. The message-passing operator within the MPNN framework is as follows:
\begin{equation} \label{eq - mpnn}
\begin{aligned}
    \mathbf{f}_i^{(k)} = \phi^v\left[\mathbf{f}_i^{(k-1)}, \oplus_{j \in \mathcal{N}(i)} \phi^e \left[\mathbf{f}_i^{(k-1)}, \mathbf{f}_j^{(k-1)}\right] \right],
\end{aligned}
\end{equation}
where $\phi^e$ and $\phi^v$ are multilayer perceptron (MLP)-based update functions, and $\oplus$ denotes a permutation invariant operator (e.g., sum, mean, max). $\mathcal{N}(i)$ denotes the neighborhood of the given node \cite{fey2019fast}. Within the SIMPNet framework, one layer of message passing is performed using eq. \ref{eq - mpnn}, considering $sum$ as the $\oplus$ operator, as follows:
\begin{equation} \label{eq - simpnetmpnn}
\begin{aligned}
    \mathbf{f}_i = \phi^v\left[\mathbf{f}_i, \sum_{j \in \mathcal{N}(i)} \phi^e \left[\mathbf{f}_i, \mathbf{f}_j\right] \right].
\end{aligned}
\end{equation}

\noindent
\textbf{Final Embedding Layer}: Finally, another deep neural network is utilized to map the output of the message passing layer to a single number representing the next timestep joint angle $\mathbf{q}_{t+1} \in \mathbb{R}^6$ within the framework of sampling-based motion planning algorithms.

\subsection{Bi-directional Planning Algorithm}
We embed our proposed sampling heuristic into the bi-directional planning algorithm proposed by \cite{qureshi2019motion} to have our proposed planning algorithm. Algorithm \ref{alg:simpnet} shows SIMPNet algorithm. Interested readers are welcome to check the work by Qureshi et al. \cite{qureshi2019motion} for a detailed description of the utilized bi-directional planning algorithm.

\begin{algorithm}[t]
\caption{SIMPNet ($\mathbf{q}_{start}, \mathbf{q}_{goal}, \mathbf{o}_{env}$)}
\label{alg:simpnet}
    \DontPrintSemicolon
    $\mathbf{q}^a \leftarrow \{\mathbf{q}_{start}\}$, $\mathbf{q}^b \leftarrow \{\mathbf{q}_{goal}\}$ \\
    $\mathbf{q}\leftarrow \emptyset$ \\
    \textit{Complete} $\leftarrow$ \textit{False} 
    
    \BlankLine
    \For{$i\leftarrow0$ to steps}{
        \Comment*[l]{forward kinematics}
        $\mathbf{x}_a \leftarrow \text{\textit{FK}}(\mathbf{q}^a(\text{end}))$, $\mathbf{x}_b \leftarrow \text{\textit{FK}}(\mathbf{q}^b(\text{end}))$ \\
        \Comment*[l]{sampling via proposed heuristic}
        $\mathbf{q}_{new}\leftarrow \text{\textit{Heuristic}}(\mathbf{q}^a(\text{end}), \mathbf{q}^b(\text{end}), \mathbf{x}_a, \mathbf{x}_b, \mathbf{o}_{env})$ \\
        $\mathbf{q}^a \leftarrow \text{\textit{Add}}(\mathbf{q}_{new})$ \\
        \textit{Complete} $\leftarrow$ \textit{Interpolation}($\mathbf{q}^a(\text{end}), \mathbf{q}^b(\text{end})$) \\
        \BlankLine
        \If{Complete}{
            $\mathbf{q}\leftarrow \text{\textit{Concatenate}}(\mathbf{q}^a, \mathbf{q}^b)$ \\
            \Return $\mathbf{q}$ \\
        }
        \Else{
            \Return $\emptyset$
        }
        \Comment*[l]{bi-directional planning}
        swap($\mathbf{q}^a, \mathbf{q}^b$) 
    }
    \BlankLine
    \If{$\mathbf{q}$}{
        \Comment*[l]{lazy path contraction}
        $\mathbf{q}\leftarrow$ \textit{PathContraction}($q$) \\
        \Comment*[l]{path collision cheching}
        \textit{CollisionFree} $\leftarrow$ \textit{CollisionChecking}($\mathbf{q}$) \\
        \BlankLine
        \If{CollisionFree}{
            \Return $\mathbf{q}$
        }\Else{
            \Comment*[l]{replanning}
            $\mathbf{q}_{new}\leftarrow$ \textit{Replanning}($\mathbf{q}$) \\
            \Comment*[l]{lazy path contraction}
            $\mathbf{q}_{new} \leftarrow$ \textit{PathContraction}($\mathbf{q}_{new}$) \\
            \Comment*[l]{path collision cheching}
            \textit{CollisionFree} $\leftarrow$ \textit{CollisionChecking}($\mathbf{q}_{new}$) \\
            \BlankLine
            \If{CollisionFree}{
                \Return $\mathbf{q}_{new}$
            }
        }
        \Return $\emptyset$
    }
\end{algorithm}%

\section{Results and Discussion} \label{sec: results and discussion}
In this section, we detail the implementation of our planner and compare its performance with state-of-the-art motion planning algorithms. The framework is developed using the PyTorch framework \cite{paszke2017automatic}, and the simulations were conducted on a computer running Linux OS, equipped with a 2.6 GHz Intel i7-1355U CPU, and a 16 GB RAM. 

\subsection{Data Collection}
\begin{figure}[htbp] 
\centering
\includegraphics[width=3.0 in]{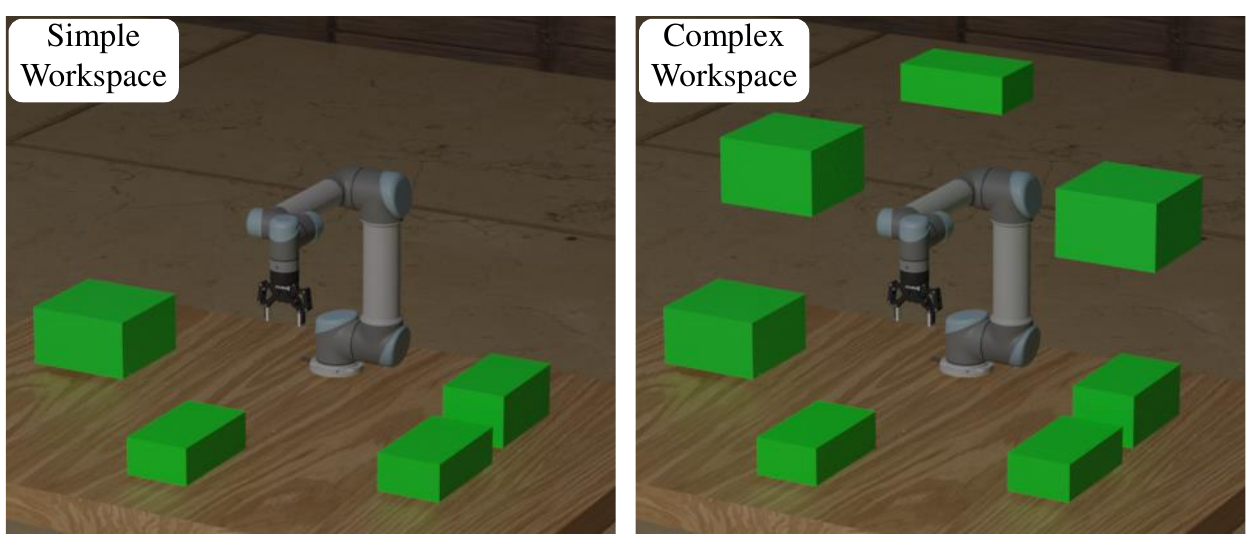}
\caption{Type of workspaces used for training and evaluation of SIMPNet.}
\label{fig_4- envs}
\end{figure}

We simulated the planning environment using MoveIt! \cite{coleman2014reducing}, and collected data using RRT* from Open Motion Planning Library (OMPL) \cite{sucan2012open}. Two types of planning environments, complex and simple, were considered. For each environment, 10 different workspaces were considered. Figure \ref{fig_4- envs} displays examples of these environments. In each workspace, 1000 collision-free paths are collected, and combined together for training. We train the network for 10,000 paths for simple and complex environments respectively. 

For testing, we evaluated the network in both seen and unseen environments from simple and complex workspaces. The seen workspace is selected from the training workspaces, while the unseen environment has the same number of obstacles with different shapes and locations compared to the workspaces used for training. 200 start-goal configurations are generated for each environment. Collision checking was performed utilizing Flexible Collision Library (FCL) \cite{pan2012fcl} within the MoveIt! framework.

\begin{figure*}[htbp] 
\centering
\includegraphics[width=\linewidth, trim={0cm, 4cm, 0.5cm, 0cm}, clip]{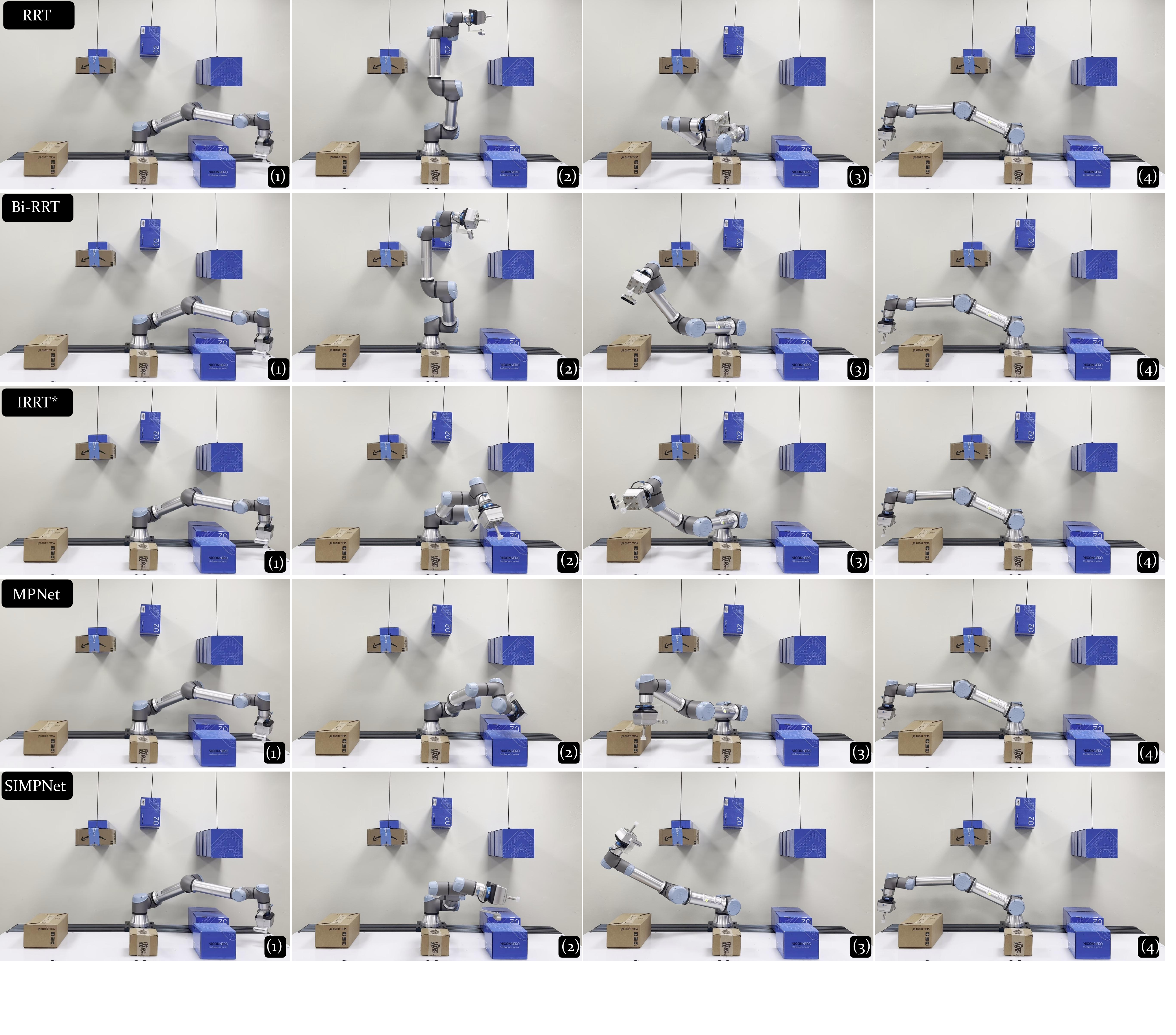}
\caption{Experimental evaluation and comparison of planned paths by SIMPNet and some baseline planners in a complex environment: Bi-RRT (Bi-directional RRT), IRRT* (Informed-RRT*), and MPNet (Motion Planning Networks) \cite{qureshi2019motion}. RRT completes the path in $32$ seconds with a planning cost of $9.32$. Bi-RRT plans in $9.44$ seconds with a planning cost of $9.02$. IRRT* takes $500$ seconds, achieving a cost of $4.01$. MPNet completes the path in $218$ seconds with a planning cost of $11.3$. SIMPNet plans the path in $5.81$ seconds with a planning cost of $8.08$. Please note that the planning cost for a path $\{\mathbf{q}_1, \mathbf{q}_2, ..., \mathbf{q}_n\}$ is calculated as $\sum_{i=0}^{n-1}||\mathbf{q}_{i+1} - \mathbf{q}_i||_2$, where $\mathbf{q}_i$ represents the robotic manipulator's configuration state, and the shortest path in the configuration space doesn't necessarily translate into the workspace. Experimental videos are available via this \href{https://zh.engr.tamu.edu/wp-content/uploads/sites/310/2024/12/Revised-SIMPNet-II-Demo.mp4}{link}.}
\label{fig_4- simpnetvsrrt}
\end{figure*}

\subsection{Network Training}
We utilize the SIMPNet sampling heuristic network parameterized by $\mathbf{\theta}$ for planning. Let $Q = \{\mathbf{q}_1, ..., \mathbf{q}_N\}$ be the training dataset where $\mathbf{q} = [\mathbf{q}_0, ..., \mathbf{q}_t, ..., \mathbf{q}_T]$ is a path planned by the oracle planner. The loss function is defined as:
\begin{equation} \label{eq - lossfn}
\begin{aligned}
    \mathcal{L}_{\text{SIMPNet}} = \frac{1}{N_P}\sum_{p=1}^{N_p}\sum_{t=0}^{T_p - 1} ||\mathbf{q}_{p, t+1} - \hat{\mathbf{q}}_{p, t+1}||^2,
\end{aligned}
\end{equation}
where $N_p$ is the batch size.

\subsection{Baselines and Metrics}
\textbf{Algorithms and Baselines.} We assessed the performance of SIMPNet by comparing it with several state-of-the-art motion planning algorithms. We chose Bi-directional RRT \cite{kuffner2000rrt} and Informed-RRT* \cite{gammell2014informed}, as baseline sampling-based motion planners. Bi-directional RRT utilizes a uniform sampling heuristic, and Informed-RRT* employs a hand-crafted sampling heuristic for sampling within the robotic manipulator's configuration space. In addition, we chose two recent deep learning-based sampling heuristics called MPNet \cite{qureshi2019motion}, and KG-Planner \cite{10576733}. Given that KG-Planner is a deterministic motion planner with limited application in simple environments, for fair comparison we made its structure stochastic. These deep learning-based planners are selected to demonstrate the effectiveness of graph construction and attention mechanisms in informed sample generation. All deep learning-based sampling heuristics used the same workspace embedding network and number of replanning attempts. Also, a Python implementation of all the planners was leveraged for comparison purposes. Figure \ref{fig_4- simpnetvsrrt} shows the performance of SIMPNet compared to other baseline motion planners, in a planning instance in a complex environment. 

\textbf{Metrics.} We employed three commonly used metrics for evaluating the performance of SIMPNet against baseline motion planners: \textit{planning time}, \textit{planning cost}, and \textit{success rate}. \textit{Planning time} ``T'' refers to the average planning time the planner takes in each environment. \textit{Planning cost} ``C'' measures the length of the successfully planned paths in the configuration space. \textit{Success rate} ``Success'' represents the percentage of successfully planned paths.

\subsection{Randomness via Dropout}
We incorporate Dropout \cite{srivastava2014dropout} within the structure of the sampling heuristic of SIMPNet, and also applied it to two deep learning-based motion planning baselines for sample generation \cite{qureshi2020motion}. This introduction of randomness implies that each planning attempt by these planners could result in a different path between any given start and goal configuration. This property, the defining feature of classical sampling-based motion planning algorithms, contributes to the generalizability of these sampling heuristics to unseen environments.

\subsection{SIMPNet Performance in a Simple Environment}
We evaluate the performance of SIMPNet, and all the baselines in a simple environment in Table \ref{tab: simpnet-performance} (see an example of a simple environment in Figure \ref{fig_4- envs}). Since classical planners used in this study (e.g., Bi-RRT and Informed-RRT*) lack inherent termination conditions, we set the planning time for these planners to match the average planning time of our proposed planner. Unlike Informed-RRT*, Bi-RRT is not a sub-optimal planner. However, it is crucial to set a termination criterion to prevent infinite searches for feasible paths.
\setlength{\tabcolsep}{2pt}
\begin{table*}[htbp]
\begin{center}
\captionof{table}{Planning performance of SIMPNet and baseline planners across all environments. Light grey is employed to distinguish the performance of deep learning-based planners from that of classical sampling-based planners. ``Bi-RRT'' refers to Bi-directional RRT, ``IRRT*'' denotes informed RRT*, ``T'' denotes \textit{planning time}, ``C'' denotes \textit{planning cost}, and ``S'' refers to \textit{success rate}. $\downarrow$ indicates lower is better, and $\uparrow$ indicates higher is better. Please note that the discrepancy in planning time between IRRT* and our planner is due to the fact that the termination time for IRRT* is set at the start of the planning algorithm.}

\label{tab: simpnet-performance}

\hspace{-0.2cm}
\begin{tabular*}{\linewidth}{@{}l
@{\extracolsep{\fill}}
c c c c c c c@{\extracolsep{\fill}} c
c c c c c c}
\toprule
\phantom{Var.} &  
\multicolumn{6}{c}{Simple Environment} && \multicolumn{6}{c}{Complex Environment}\\
\cmidrule{2-7}
\cmidrule{9-14}
&\multicolumn{3}{c}{Seen}&\multicolumn{3}{c}{Unseen}&&\multicolumn{3}{c}{Seen}&\multicolumn{3}{c}{Unseen} \\
\cmidrule{2-4}
\cmidrule{5-7}
\cmidrule{9-11}
\cmidrule{12-14}
& {T $[s]\downarrow$} & {C $\downarrow$} & {S $[\%]\uparrow$} & {T $[s]\downarrow$} & {C $\downarrow$} & {S $[\%]\uparrow$} & {} & {T $[s]\downarrow$} & {C $\downarrow$} & {S $[\%]\uparrow$}& {T $[s]\downarrow$} & {C $\downarrow$} & {S $[\%]\uparrow$} \\
\toprule
\scriptsize{Bi-RRT}&\scriptsize{$1.1\pm .5$}&\scriptsize{$10\pm 3.2$}&\scriptsize{\textbf{97\%}}&\scriptsize{$1.22\pm .7$}&\scriptsize{$10.1\pm 3.5$}&\scriptsize{\textbf{97\%}}&&\scriptsize{$\mathbf{3.2 \pm 1.8}$}&\scriptsize{$8.6 \pm 4.8$}&\scriptsize{\textbf{75\%}}&\scriptsize{$\mathbf{4.9 \pm 3.1}$}&\scriptsize{$9.9 \pm 5.9$}&\scriptsize{\textbf{68\%}} \\
\midrule
\scriptsize{IRRT}*&\scriptsize{$1.2\pm .2$}&\scriptsize{$\mathbf{5.2\pm .9}$}&\scriptsize{77\%}&\scriptsize{$1.3\pm .2$}&\scriptsize{$\mathbf{5.3\pm 1.1}$}&\scriptsize{84\%}&&\scriptsize{$7.5 \pm .0$}&\scriptsize{$\mathbf{4.7 \pm 1.5}$}&\scriptsize{45\%}&\scriptsize{$6.4 \pm .1$}&\scriptsize{$\mathbf{4 \pm 1.8}$}&\scriptsize{34\%} \\
\toprule
\scriptsize{MPNet \cite{qureshi2019motion}}&\scriptsize{$.95\pm .4$}&\scriptsize{$5.9\pm 2.1$}&\scriptsize{94\%}&\scriptsize{$.94\pm .6$}&\scriptsize{$5.7\pm .8$}&\scriptsize{92\%}&&\scriptsize{$7.1 \pm 4.1$}&\scriptsize{$4.8 \pm 1.5$}&\scriptsize{47\%}&\scriptsize{$20.1 \pm 9.6$}&\scriptsize{$4.4 \pm 1.9$}&\scriptsize{37\%} \\
\midrule
\scriptsize{KG-Planner \cite{10576733}}&\scriptsize{$\mathbf{.80\pm .4}$}&\scriptsize{$5.8\pm 1.7$}&\scriptsize{93\%}&\scriptsize{$.91\pm .6$}&\scriptsize{$5.8\pm 1.9$}&\scriptsize{\textbf{97\%}}&&\scriptsize{$4. \pm 3.8$}&\scriptsize{$5.7 \pm 2.3$}&\scriptsize{60\%}&\scriptsize{$6.8 \pm 3.1$}&\scriptsize{$5.7 \pm 2.9$}&\scriptsize{38\%} \\
\bottomrule
\scriptsize{SIMPNet \textbf{(ours)}}&\scriptsize{\underline{$.82 \pm 1.$}}&\scriptsize{\underline{$5.79 \pm 2.1$}}&\scriptsize{\underline{95\%}}&\scriptsize{\underline{$\mathbf{.67 \pm .9}$}}&\scriptsize{\underline{$5.9 \pm 2$}}&\scriptsize{\underline{\textbf{97\%}}}&&\scriptsize{\underline{$7.4 \pm 4.8$}}&\scriptsize{\underline{$6.3 \pm 2.9$}}&\scriptsize{\underline{72\%}}&\scriptsize{\underline{$6.3 \pm 3.3$}}&\scriptsize{\underline{$5.9 \pm 3$}}&\scriptsize{\underline{65\%}} \\
\bottomrule
\end{tabular*}
\end{center}
\end{table*}

\begin{figure}[htbp] 
\centering
\includegraphics[width=3.2 in]{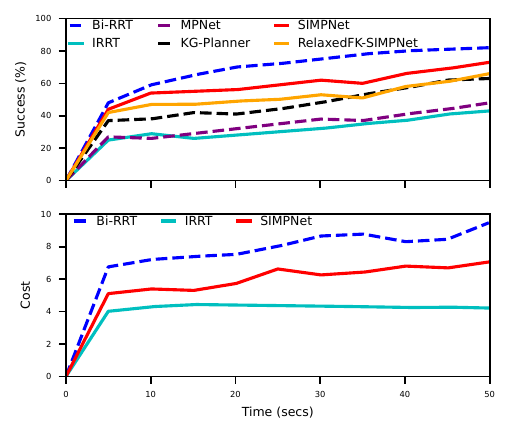}
\caption{Planning results in an ``unseen'' complex environment: comparison of SIMPNet with the benchmark planners.}
\label{fig_4- time_success}
\vspace{-0.1cm}
\end{figure}

The results show that Bi-RRT has the highest success rate among all the planners. Its inherent bi-directional heuristic contributes to its efficiency, making it fast. However, this planner compromises planning cost for higher planning time and success rate \cite{gammell2015batch}. Informed-RRT* rewires the constructed graph during planning to find the sub-optimal path, resulting in the lowest planning cost among the utilized planners, while trading off success rate and planning time. Among the deep learning-based planners, the proposed planner performs comparably similarly to others. This can be attributed to the fact that the simple workspace encoding by MPNet, and the naive graph representation by KG-Planner are efficient for planning in simple environments. Also, the use of Dropout during planning, helps all the deep learning-based planners generalize effectively to unseen environments.

\subsection{SIMPNet Performance in a Complex Environment}
We evaluate the performance of SIMPNet, and all the baselines in a complex environment in Table \ref{tab: simpnet-performance} (see an example of a complex environment in Figure \ref{fig_4- envs}). We also set the planning time of Bi-RRT and Informed-RRT* as the average planning time of the SIMPNet algorithm. In this environment, Bi-RRT maintains the highest success rate and planning time across seen and unseen environments, benefiting from its bi-directional module. However, it does so at the expense of increased planning costs. Similarly, Informed-RRT*, with its inherent rewiring heuristic, achieves the lowest planning cost, while trading off success rate and planning time. Also, Figure \ref{fig_4- time_success} illustrates the success of the proposed planner, SIMPNet, compared to benchmark planners in an unseen complex environment.

SIMPNet achieves a higher success rate and lower planning time compared to other baseline deep learning-based planners. This performance can be attributed to the proposed approach of encoding the kinematic chain of the robotic manipulator as a graph. MPNet concatenates the workspace encoding with the planning problem information, ignoring the spatial correlation inherent in the planning problem. The results in Table \ref{tab: simpnet-performance} demonstrate the effectiveness of using graph representation over the approach adopted by MPNet.

Unlike other baseline deep learning-based planners such as KG-Planner, which directly integrates the configuration space and workspace into the sampling heuristic network using a simple graph representation, SIMPNet employs a cross-attention mechanism to condition the planning problem on workspace encoding. The results in Table \ref{tab: simpnet-performance} illustrate the benefits of using the attention mechanism in our framework compared to KG-Planner.

\subsection{Ablation Study: Forward Kinematics Relaxed-SIMPNet (RelaxedFK-SIMPNet)}
Although the original graph was constructed to implicitly represent the kinematic structure of the robotic manipulator in the configuration space, node features also contain some information from the workspace, such as joint positions, and goal positions. However, using forward kinematics in the planning and replanning phase of the proposed planner results in a trade-off between planning time and success rate. Here we are considering constructing a modified graph based solely on information from the configuration space. The newly constructed graph retains the same nodes and edges as the original graph but features the following node attributes ($\mathbf{f}_i \in \mathbb{R}^3$).
\begin{equation} \label{eq - node_featuresnofk}
\begin{aligned}
    \mathbf{f}_i = [q_t, q_{goal}, |q_t - q_{goal}|],
\end{aligned}
\end{equation}
where $q_t~\in \mathbb{R}$, and $q_{goal}~\in \mathbb{R}$ are current time-step and goal joint angles within the configuration space, respectively. The performance of RelaxedFK is compared against SIMPNet across all the environments. Table \ref{tab: planning performance - ablation-simpnetvsrelaxedFKsimpnet} showcases the performance metrics of these planners.

\begin{table}[htbp]
\centering
\caption{Comparison of planning performance between SIMPNet and RelaxedFK-SIMPNet across all environments. ``SMP'' denotes SIMPNet, ``RFK-SMP'' denotes RelaxedFK-SIMPNet.}
\label{tab: planning performance - ablation-simpnetvsrelaxedFKsimpnet}
\renewcommand{\arraystretch}{1.5} 
\begin{tabular}{@{}p{0.09\linewidth}p{0.07\linewidth}|p{0.1\linewidth}p{0.2\linewidth}|p{0.1\linewidth}p{0.2\linewidth}@{}}
\toprule
\multicolumn{2}{c}{}&\multicolumn{2}{c}{Simple Environment}&\multicolumn{2}{c}{Complex Environment} \\
\toprule
&\multicolumn{1}{c}{Planner}&\multicolumn{1}{c}{T $[s]\downarrow$}&\multicolumn{1}{c}{S $[\%]\uparrow$}&\multicolumn{1}{c}{T $[s]\downarrow$}&\multicolumn{1}{c}{S $[\%]\uparrow$} \\
\toprule
\multirow{2}{*}{\scriptsize{Seen}} & \multicolumn{1}{c|}{\scriptsize{SMP}} & \multicolumn{1}{c}{\scriptsize{$\mathbf{.82 \pm 1}$}} & \multicolumn{1}{c|}{\scriptsize{95\%}} &  \multicolumn{1}{c}{\scriptsize{$7.4 \pm 4.8$}}& \multicolumn{1}{c}{\scriptsize{\textbf{72\%}}} \\\hhline{~-||--||--}
                      & \multicolumn{1}{c|}{\scriptsize{RFK-SMP}} & \multicolumn{1}{c}{\scriptsize{$.68 \pm .7$}} & \multicolumn{1}{c|}{\textbf{\scriptsize{96\%}}} &  \multicolumn{1}{c}{\scriptsize{$\mathbf{2.6 \pm 1.9}$}}& \multicolumn{1}{c}{\scriptsize{63\%}} \\
\toprule
\multirow{2}{*}{\scriptsize{Unseen}} & \multicolumn{1}{c|}{\scriptsize{SMP}} & \multicolumn{1}{c}{\scriptsize{$\mathbf{.67 \pm .9}$}} & \multicolumn{1}{c|}{\scriptsize{97\%}} &  \multicolumn{1}{c}{\scriptsize{$\mathbf{6.3 \pm 3.3}$}}& \multicolumn{1}{c}{\scriptsize{\textbf{65\%}}} \\\hhline{~-||--||--}
                      & \multicolumn{1}{c|}{\scriptsize{RFK-SMP}} & \multicolumn{1}{c}{\scriptsize{$0.7 \pm .2$}} & \multicolumn{1}{c|}{\scriptsize{\textbf{98\%}}} &  \multicolumn{1}{c}{\scriptsize{$6.9 \pm 3.1$}}& \multicolumn{1}{c}{\scriptsize{54\%}}\\
\bottomrule
\end{tabular}
\end{table}

The results suggest that RelaxedFK-SIMPNet trades off success rate for reduced planning time. This property is desirable in simple planning environments, as it is highly probable that the proposed planner can find a path without incorporating forward kinematics information, The slight differences observed between SIMPNet and RelaxedFK-SIMPNet are primarily driven by the multi-modality nature of the informed sampling process. However, for complex environments, the results indicate that including forward kinematics enhances the success rate of the motion planner.

\subsection{Ablation Study: Validating SIMPNet for Various Number of Obstacles}
Although the workspace embedding for this framework is a fixed vector, SIMPNet can be validated by fewer obstacles compared to training workspaces by padding the workspace embedding with zeros during validation. This adjustment maintains the required fixed-size embedding for informed sampling. Table \ref{tab: planning performance - vary-obs-number} details the results.

\begin{table}[htbp]
\caption{Comparison of SIMPNet's planning performance within workspaces with a different number of obstacles.}
\label{tab: planning performance - vary-obs-number}
\renewcommand{\arraystretch}{1.5}
\centering
\begin{tabular}{@{}p{0.25\linewidth}|p{0.1\linewidth}p{0.1\linewidth}p{0.2\linewidth}@{}}
\toprule
No. of Obstacles&\multicolumn{1}{c}{T $[s]\downarrow$} &\multicolumn{1}{c}{C $\downarrow$} &\multicolumn{1}{c}{S $[\%]\uparrow$}\\
\toprule
\multicolumn{1}{c|}{\scriptsize{4}}&\multicolumn{1}{c}{\scriptsize{$.76 \pm .4$}}&\multicolumn{1}{c}{\scriptsize{$4.9 \pm 1.6$}}&\multicolumn{1}{c}{\scriptsize{90\%}} \\
\midrule
\multicolumn{1}{c|}{\scriptsize{5}}&\multicolumn{1}{c}{\scriptsize{$2.8 \pm 1.3$}}&\multicolumn{1}{c}{\scriptsize{$6.1 \pm 3.52$}}&\multicolumn{1}{c}{\scriptsize{70\%}} \\
\midrule
\multicolumn{1}{c|}{\scriptsize{6}}&\multicolumn{1}{c}{\scriptsize{$3.5 \pm 1.5$}}&\multicolumn{1}{c}{\scriptsize{$6.2 \pm 3.55$}}&\multicolumn{1}{c}{\scriptsize{62\%}} \\
\midrule
\multicolumn{1}{c|}{\scriptsize{7}}&\multicolumn{1}{c}{\scriptsize{$8.1 \pm 6.1$}}&\multicolumn{1}{c}{\scriptsize{$7.6 \pm 4.14$}}&\multicolumn{1}{c}{\scriptsize{55\%}} \\
\bottomrule
\end{tabular}
\end{table}

The results presented in Table \ref{tab: planning performance - vary-obs-number} indicate that simplifying the workspace will increase the planning efficiency of the proposed planner due to the reduced number of collision checking.
\section{Conclusion and Limitations} \label{sec: conclusion}
In this paper, we introduced SIMPNet, a bi-directional motion planner that utilizes an informed deep learning-based sampling heuristic. We constructed a graph representing the kinematic chain of the robotic manipulator within the configuration space and employed a cross-attention mechanism to integrate information from the 3D workspace to the robotic manipulator's 6D configuration space. The proposed sampling heuristic within SIMPNet was trained via supervised learning on optimal paths generated by an oracle planner. Additionally, We used Dropout within during inference to introduce stochastic behavior in the SIMPNet architecture. 

Our results highlight the advantage of using graphs and graph neural networks. These tools are highly effective for retaining and leveraging spatial relationships inherent in the configuration space and kinematic structure of the robotic manipulators. This capability enhances informed sampling within the sampling-based motion planning algorithms. Additionally, the attention mechanism was effectively employed to integrate workspace information into the configuration space, facilitating highly informed sample generation.

SIMPNet has a limitation in that it assumes the shape and location of obstacles are known, which is often not true in real-time planning scenarios. One solution could be to integrate a perception module with SIMPNet to accurately encode the workspace. Another issue is that SIMPNet is mainly suited for static environments. To adapt it to dynamic environments, the bi-directional planner \cite{qureshi2019motion} needs to be modified for efficient replanning.
\bibliographystyle{IEEEtran}
\bibliography{ref} 

\end{document}